\documentclass[letterpaper, 10 pt, conference]{ieeeconf}  % Comment this line out if you need a4paper
\IEEEoverridecommandlockouts                              % This command is only needed if 
\usepackage{url}
\usepackage{amsmath,amssymb}
\usepackage{graphicx,color}
\usepackage{verbatim}
\usepackage{gensymb}
\graphicspath{{figures/}}
\newcommand{\ba}{\mathbf{a}}

\newcommand{\bo}{\mathbf{o}}

\usepackage{caption}
\def\tablename{Table}

\title{\LARGE \bf
Deep Imitation Learning of Sequential Fabric Smoothing\\ From an Algorithmic Supervisor
}

\author{Daniel Seita$^1$, Aditya Ganapathi$^{1}$, Ryan Hoque$^{1}$, Minho Hwang$^1$, Edward Cen$^1$, \\  Ajay Kumar Tanwani$^1$, Ashwin Balakrishna$^1$, Brijen Thananjeyan$^1$, Jeffrey Ichnowski$^1$, \\ Nawid Jamali$^2$, Katsu Yamane$^2$, Soshi Iba$^2$, John Canny$^1$, Ken Goldberg$^{1}$% <-this % stops a space
\thanks{$^{1}$AUTOLAB at the University of California, Berkeley, USA.}%
\thanks{$^{2}$Honda Research Institute, USA.}%
\thanks{Correspondence to {\tt\small seita@berkeley.edu}}%
}
\begin{document}

\maketitle
\thispagestyle{empty}
\pagestyle{empty}

%%%%%%%%%%%%%%%%%%%%%%%%%%%%%%%%%%%%%%%%%%%%%%%%%%%%%%%%%%%%%%%%%%%%%%%%%%%%%%%%
\begin{abstract}
Sequential pulling policies to flatten and smooth fabrics have applications from surgery to manufacturing to home tasks such as bed making and folding clothes. Due to the complexity of fabric states and dynamics, we apply deep imitation learning to learn policies that, given color (RGB), depth (D), or combined color-depth (RGBD) images of a rectangular fabric sample, estimate pick points and pull vectors to spread the fabric to maximize coverage. To generate data, we develop a fabric simulator and an algorithmic supervisor that has access to complete state information. We train policies in simulation using domain randomization and dataset aggregation (DAgger) on three tiers of difficulty in the initial randomized configuration. We present results comparing five baseline policies to learned policies and report systematic comparisons of RGB vs D vs RGBD images as inputs. In simulation, learned policies achieve comparable or superior performance to analytic baselines. In 180 physical experiments with the da Vinci Research Kit (dVRK) surgical robot, RGBD policies trained in simulation attain coverage of 83\% to 95\% depending on difficulty tier, suggesting that effective fabric smoothing policies can be learned from an algorithmic supervisor and that depth sensing is a valuable addition to color alone. Supplementary material is available at \url{https://sites.google.com/view/fabric-smoothing}. 
\end{abstract}
%%%%%%%%%%%%%%%%%%%%%%%%%%%%%%%%%%%%%%%%%%%%%%%%%%%%%%%%%%%%%%%%%%%%%%%%%%%%%%%%

\section{Introduction}\label{sec:intro}

Robot manipulation of fabric has applications in senior care and dressing assistance~\cite{deep_dressing_2018,ra-l_dressing_2018,personalized_dressing_2016}, sewing~\cite{sewing_2012}, ironing~\cite{ironing_2016}, laundry folding~\cite{folding_iros_2015,laundry2012,shibata2012trajectory,folding_2017}, fabric upholstery manufacturing~\cite{10.1115/1.3185859, Torgerson1987VisionGR}, and handling gauze in robotic surgery~\cite{thananjeyan2017multilateral}. However, fabric manipulation is challenging due to its infinite dimensional configuration space and unknown dynamics.

%Fabric manipulation is challenging due to the infinite dimensional configuration space of fabric and unknown dynamics.  Robot manipulation of fabric has applications in senior care and dressing assistance~\cite{deep_dressing_2018,ra-l_dressing_2018,personalized_dressing_2016}, sewing~\cite{sewing_2012}, ironing~\cite{ironing_2016}, folding of towels and laundry~\cite{folding_iros_2015,laundry2012,shibata2012trajectory,folding_2017}, fabric manufacturing~\cite{10.1115/1.3185859, Torgerson1987VisionGR}, and handling gauze in robotic surgery~\cite{thananjeyan2017multilateral}.

\begin{figure}[t]
\center
\includegraphics[width=0.48\textwidth]{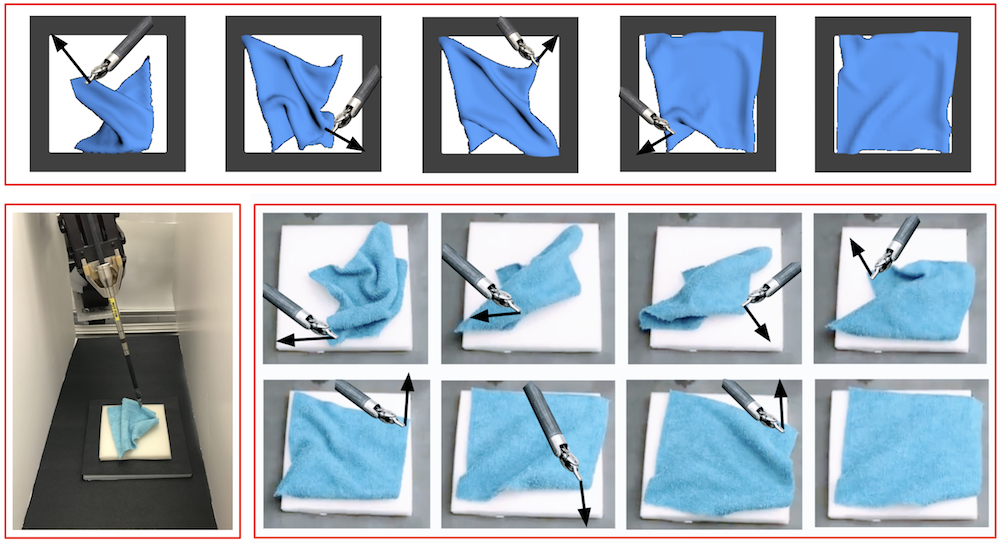}
\caption{
\small
Learned policies executed in simulation and with a physical da Vinci surgical robot, with actions indicated from the overlaid arrows. Policies are learned in simulation using DAgger with an algorithmic supervisor that has full state information, using structured domain randomization with color and/or depth images. The 4-action smoothing episode in simulation (top) increases coverage from 43\% to 95\%. The 7-action episode on the physical da Vinci robot (bottom) increases coverage from 49\% to 92\%.
}
\vspace*{-10pt}
\label{fig:teaser}
\end{figure}

% Motivate the task / problem better?
We consider the task of transforming fabric from a rumpled and highly disordered starting configuration to a smooth configuration via a series of grasp and pull actions. We explore a deep imitation learning approach based on a Finite Element Method (FEM) fabric simulator with an algorithmic supervisor and use DAgger~\cite{ross2011reduction} to train policies.
%We use a corner pulling supervisor that has access to the internal state. The supervisor sequentially selects a fabric corner (even if occluded) and pulls it to a known target to smooth the fabric.
Using color and camera domain randomization~\cite{cad2rl,domain_randomization}, learned policies are evaluated in simulation and in physical experiments with the da Vinci Research Kit (dVRK) surgical robot~\cite{dvrk2014}. Figure~\ref{fig:teaser} shows examples of learned smoothing episodes in simulation and the physical robot.

This paper contributes: (1) an open-source simulation environment and dataset for evaluation of fabric smoothing with three difficulty tiers of initial fabric state complexity, (2) deep imitation learning of fabric smoothing policies from an algorithmic supervisor using a sequence of pick and pull actions, and (3) transfer to physical experiments on a da Vinci surgical robot with comparisons of coverage performance using color (RGB), depth (D), or RGBD input images.

%\bnote{This paper contributes: (1) a novel formulation of fabric smoothing in terms of a sequence of pick and pull actions, (2) a simulation environment for data generation and evaluation of fabric smoothing with three difficulty tiers of initial state complexity in terms of coverage and visible corners, (3) deep imitation learning of fabric smoothing policies in simulation, and (4) physical experiments of the simulation policy transferred to a da Vinci surgical robot which compare coverage performance using color vs. depth input images across all tiers.}

\section{Related Work}\label{sec:rw}

Well-known research on robotic fabric manipulation~\cite{grasp_centered_survey_2019,manip_deformable_survey_2018} uses bilateral robots and gravity to expose corners. Osawa et al.~\cite{osawa_2007} proposed a method of iteratively re-grasping the lowest hanging point of a fabric to flatten and classify fabrics. Subsequently, Kita et al.~\cite{kita_2009_iros,kita_2009_icra} used a deformable object model to simulate fabric suspended in the air, allowing the second gripper to grasp at a desired point. Follow-up work generalized to a wider variety of initial configurations of new fabrics.  In particular, Maitin-Shepard~et~al.~\cite{maitin2010cloth}, Cusumano-Towner~et~al.~\cite{cusumano2011bringing}, and Doumanoglou~et~al.~\cite{unfolding_rf_2014} identified and tensioned corners to fold laundry or to bring clothing to desired positions. These methods rely on gravity to reveal corners of the fabric. We consider the setting where a single armed robot adjusts a fabric strewn across a surface without lifting it entirely in midair, which is better suited for larger fabrics or when robots have a limited range of motion.

\subsection{Reinforcement Learning for Fabric Manipulation}

Reinforcement Learning (RL)~\cite{Sutton_2018} is a promising method for training policies that can manipulate highly deformable objects. In RL applications for folding, Matas~et~al.~\cite{sim2real_deform_2018} assumed that fabric is flat, and Balaguer~et~al.~\cite{balaguer2011combining} began with fabric gripped in midair to loosen wrinkles. In contrast, we consider the problem of bringing fabric from a highly rumpled configuration to a flat configuration. Using model-based RL, Ebert~et~al.~\cite{visual_foresight_2018} were able to train robots to fold pants and fabric. This approach requires executing a physical robot for many thousands of actions and then training a video prediction model. In surgical robotics, Thananjeyan~et~al.~\cite{thananjeyan2017multilateral} used RL to learn a tensioning policy to cut gauze, with one arm pinching at a pick point to let the other arm cut. We focus on fabric smoothing without tensioning, and additionally consider cases where the initial fabric state may be highly rumpled and disordered. In concurrent and independent work, Jangir~et~al.~\cite{rishabh_2020} used deep reinforcement learning with demonstrations to train a policy using fabric state information for simulated dynamic folding tasks. 

\subsection{Fabric Smoothing}

In among the most relevant prior research on fabric smoothing, Willimon~et~al.~\cite{willimon_unfolding_laundry_2011} present an algorithm that pulls at eight fixed angles, and then uses a six-step stage to identify corners from depth images using the Harris Corner Detector~\cite{harris_1988}. They present experiments on three simulated trials and one physical robot trial. Sun~et~al.~\cite{heuristic_wrinkles_2014} followed up by attempting to explicitly detect and then pull at wrinkles. They measure wrinkledness as the average absolute deviation in a local pixel region for each point in a depth map of the fabric~\cite{depth_wrinkles_2012} and apply a force perpendicular to the largest wrinkle. Sun~et~al. evaluate on eight fixed, near-flat fabric starting configurations in simulation. In subsequent work, Sun~et~al.~\cite{cloth_icra_2015} improved the detection of wrinkles by using a shape classifier as proposed in Koenderink and van Doorn~\cite{koenderink1992surface}. Each point in the depth map is classified as one of nine shapes, and they use contiguous segments of certain shapes to define a wrinkle. While Sun~et~al. were able to generalize the method beyond a set of hard-coded starting states, it was only tested on nearly flat fabrics in contrast to the highly rumpled configurations we explore.

In concurrent and independent work, Wu~et~al.~\cite{lerrel_2020} trained an image-based policy for fabric smoothing in simulation using deep reinforcement learning, and then applied domain randomization to transfer it to a physical PR2 robot.

This paper extends prior work by Seita~et~al.~\cite{seita-bedmaking} that only estimated a pick point and pre-defined the pull vector. In contrast, we learn the pull vector and pick point simultaneously. Second, by developing a simulator, we generate far more training data and do not need to run a physical robot. This enables us to perform systematic experiments comparing RGB, D, and RGBD image inputs.

\section{Problem Statement}\label{sec:PS}

Given a deformable fabric and a flat fabric plane, each with the same rectangular dimensions, we consider the task of manipulating the fabric from a start state to one that maximally covers the fabric plane. We define an \emph{episode} as one instance of the fabric smoothing task.

Concretely, let $\xi_t$ be the full state of the fabric at time $t$ with positions of all its points (described in Section~\ref{ssec:fabric-sim}). Let $\bo_t \in O$ be the \emph{image observation} of the fabric at time $t$, where $O = \mathbb{R}^{H\times W \times c}$ represents the space of images with $H\times W$ pixels, and $c = 1$ channels for depth images, $c = 3$ for color images, or $c=4$ for combined color and depth (i.e., RGBD) images. Let $A$ be the set of actions the robot may take (see Section~\ref{ssec:act}). The task performance is measured with \emph{coverage} $C(\xi_t)$, or the percentage of the fabric plane covered by $\xi_t$. 

We frame this as imitation learning~\cite{argall2009survey}, where a supervisor provides data in the form of paired observations and actions $\mathcal{D} = \{(\bo_t, \ba_t)\}_{t=1}^N$. From $\mathcal{D}$, the robot's goal is to learn a policy $\pi: O \to A$ that maps an observation to an action, and executes sequentially until a coverage threshold or iteration termination threshold is reached.

\section{Fabric and Robot Simulator}\label{ssec:fabric-sim}

\begin{figure}[t]
\center
\includegraphics[width=0.48\textwidth]{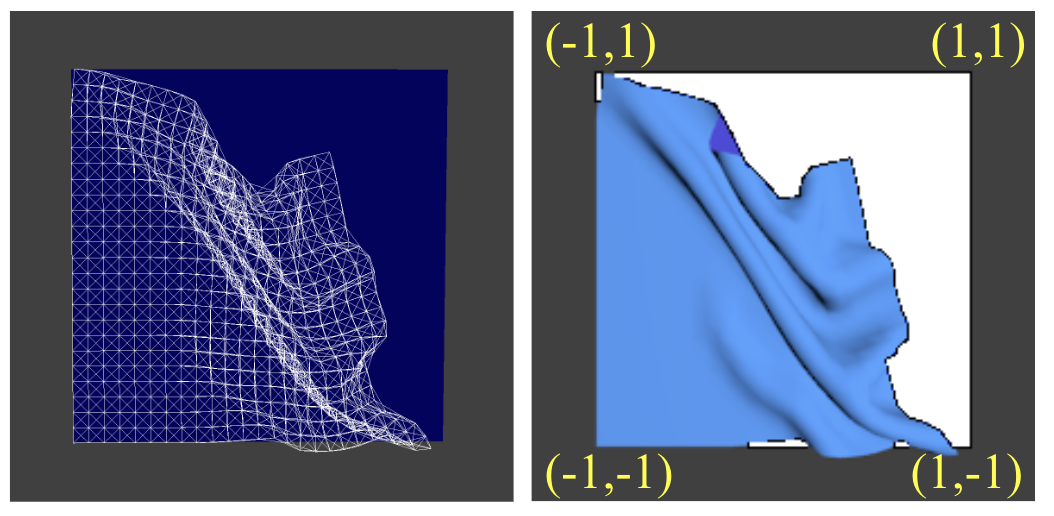}
\caption{
\small
FEM fabric simulation. Left: a wireframe rendering, showing the $25\times 25$ grid of points and the spring-mass constraints.  Right: the corresponding image with the white fabric plane, rendered using Blender and overlaid with coordinates. The coverage is 73\%, measured as the percentage of the fabric plane covered.
}
\vspace*{-10pt}
\label{fig:fabric-sim}
\end{figure}

%(\url{https://www.blender.org/}),
We implement a Finite Element Method (FEM)~\cite{bathe2006finite} fabric simulator and interface with an OpenAI gym environment design~\cite{gym}. The simulator is open source and available on the project website. Alternative fabric simulators exist, such as from Blender~\cite{blender}, which is a popular open-source computer graphics software toolkit. Since 2017, Blender has had significant improvements in physics and realism of fabrics, but these changes are only supported in Blender 2.80, which does not support headless rendering of images and therefore meant we could not run massive data collection. We downgraded to an older version of Blender, 2.79, which supports headless rendering, to create images generated from the proposed custom-built fabric simulator. MuJoCo 2.0 provides another fabric simulator, but did not support OpenAI-gym style fabric manipulation environments until concurrent work~\cite{lerrel_2020}.

The fabric (Figure~\ref{fig:fabric-sim}) is represented as a grid of $25\times 25$ point masses, connected by three types of springs~\cite{provot_1996}:

\begin{itemize}%[noitemsep]
    \item \emph{Structural}: between a point mass and the point masses to its left and above it.
    \item \emph{Shear}: between a point mass and the point masses to its diagonal upper left and diagonal upper right.
    \item \emph{Flexion}: between a point mass and the point masses two away to its left and two above it.
\end{itemize}

% Daniel: I don't know what paper to reference for this. It comes from Berkeley's CS 184 class.
%The springs modeling the bending constraints are weaker than their structural and shearing counterparts.
Each point mass is acted upon by both an external gravitational force which is calculated using Newton's Second Law and a spring correction force
\begin{equation}\label{eq:hookes}
F_s = k_s \cdot ( \|q_a - q_b\|_2 - \ell),
\end{equation}
for each of the springs representing the constraints above, where $k_s$ is a spring constant, $q_a \in \mathbb{R}^3$ and $q_b \in \mathbb{R}^3$ are positions of any two point masses connected by a spring, and $\ell$ is the default spring length. We update the point mass positions using Verlet integration~\cite{verlet_1967}. Verlet integration computes a point mass's new position at time $t + \Delta_t$, denoted with $p_{t + \Delta_t}$, as:
\begin{equation}
p_{t + \Delta_t} = p_t + v_t \Delta_t + a_t \Delta_t^2,
\end{equation}
where $p_t \in \mathbb{R}^3$ is the position, $v_t \in \mathbb{R}^3$ is the velocity, $a_t \in \mathbb{R}^3$ is the acceleration from all forces, and $\Delta_t \in \mathbb{R}$ is a timestep. Verlet integration approximates $v_t \Delta_t = p_t - p_{t - \Delta_t}$ where $p_{t - \Delta_t}$ is the position at the last time step, resulting in
\begin{equation}
p_{t + \Delta_t} = 2p_t - p_{t - \Delta_t} + a_t \Delta_t^2
\end{equation}
The simulator adds damping to simulate loss of energy due to friction, and scales down $v_t$, leading to the final update:
\begin{equation}\label{eq:final}
p_{t + \Delta_t} = p_t + (1 - d) (p_t - p_{t - \Delta_t}) + a_t\Delta_t^2
\end{equation}
where $d \in [0,1]$ is a damping term, which we tuned to 0.02 based on visually inspecting the simulator.

We apply a constraint from Provot~\cite{provot_1996} by correcting point mass positions so that spring lengths are at most 10\% greater than $\ell$ at any time. We also implement fabric-fabric collisions following~\cite{cloth-cloth-collisions} by adding a force to ``separate'' two points if they are too close.

The simulator provides access to the full fabric state $\xi_t$, which contains the exact positions of all $25\times 25$ points, but does not provide image observations $\bo_t$ which are more natural and realistic for transfer to physical robots. To obtain image observations of a given fabric state, we create a triangular mesh and render using Blender.
%(\url{https://www.blender.org/}). Blender is open-source software that can render images and simulate lighting and camera positions.

%Blender also includes its own fabric simulator, with significant improvements in physics and realism since 2017. These changes, however, are only supported in Blender 2.80, which does not support headless rendering of images and therefore meant we could not run massive data collection. We downgraded to an older version of Blender, 2.79, which supports headless rendering, to create images generated from the proposed custom-built fabric simulator. MuJoCo 2.0 provides another fabric simulator, but did not support OpenAI-gym style fabric manipulation environments until concurrent work~\cite{lerrel_2020}.

\subsection{Actions}\label{ssec:act}

We define an action at time $t$ as a 4D vector which includes the pick point $(x_t,y_t)$ represented as the coordinate over the fabric plane to grasp, along with the pull direction. The simulator implements actions by grasping the top layer of the fabric at the pick point. If there is no fabric at $(x_t,y_t)$, the grasp misses the fabric. After grasping, the simulator pulls the picked point upwards and towards direction $\Delta x_t \in [-1,1]$ and $\Delta y_t \in [-1,1]$, deltas in the $x$ and $y$ direction of the fabric plane.
%\footnote{We tried an alternative parameterization based on length and angle, but the angle is problematic at the discontinuity of $-\pi$ and $\pi$.}
In summary, actions $\ba_t \in A$ are defined as:
\begin{equation}
    \ba_t  = \langle x_t, y_t, \Delta x_t, \Delta y_t\rangle
\end{equation}
representing the pick point coordinates $(x_t,y_t)$ and the pull vector ($\Delta x_t, \Delta y_t)$ relative to the the pick point.

\subsection{Starting State Distributions}\label{ssec:starting-state}

% See: https://github.com/BerkeleyAutomation/gym-cloth/issues/28
% And for the Google drive figure, I use this file: start_distribution_04_rgb_and_depth
\begin{figure}[t]
\center
\includegraphics[width=0.48\textwidth]{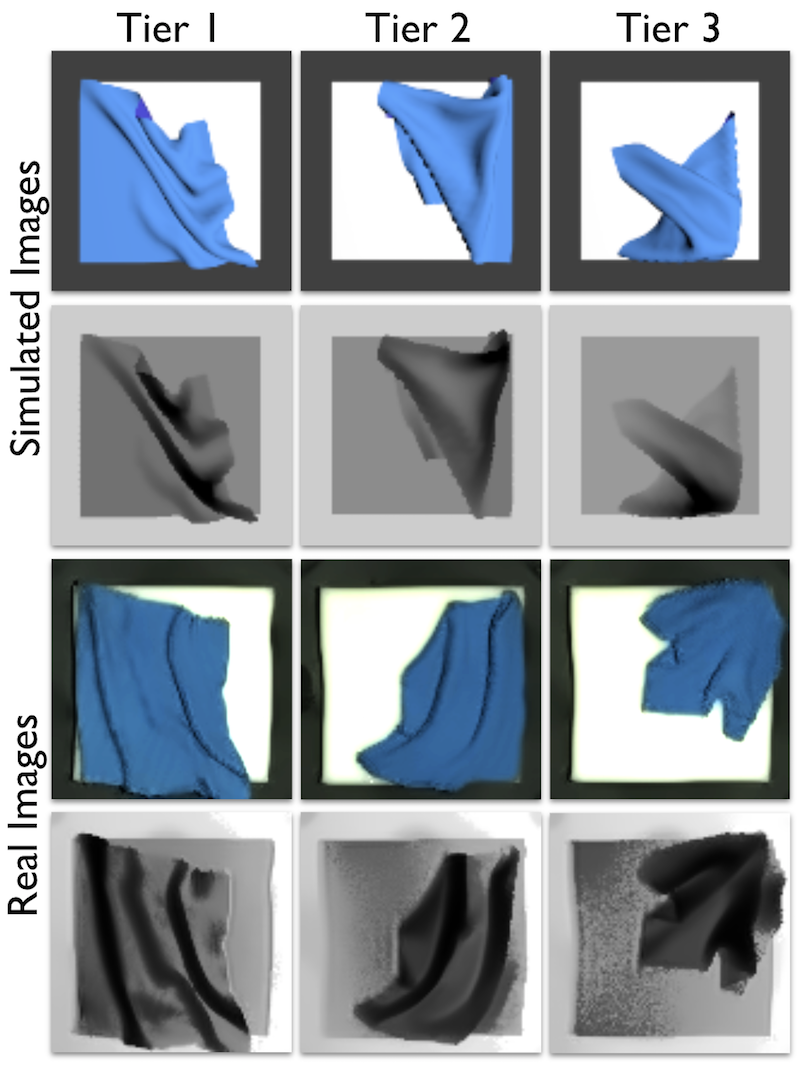}
\caption{
\small
Initial fabric configurations drawn from the distributions specified in Section~\ref{ssec:starting-state}, with tiers grouped by columns. The first two rows show representative simulated color (RGB) and depth (D) images, respectively, while the last two rows show examples of real images from a mounted Zivid One Plus camera, after smoothing and de-noising, which we then pass as input to a neural network policy. Domain randomization is not applied on the simulated images shown here.
}
\vspace*{-10pt}
\label{fig:fabric-start-states}
\end{figure}

The performance of a smoothing policy, or more generally any fabric manipulation policy, depends heavily on the distribution of starting fabric states. We categorize episodes as belonging to one of three custom difficulty tiers. For each tier, we randomize the starting state of each episode. The starting states, as well as their average coverage based on 2000 simulations, are generated as follows:

\begin{itemize}
    \item \textbf{Tier 1, $78.3\pm 6.9$\% Coverage (High)}: starting from a flat fabric, we make two short, random pulls to slightly perturb the fabric. All fabric corners remain visible.
    \item \textbf{Tier 2, $57.6 \pm 6.1$\% Coverage (Medium)}: we let the fabric drop from midair on one side of the fabric plane, perform one random grasp and pull across the plane, and then do a second grasp and pull to cover one of the two fabric corners furthest from its plane target.
    \item \textbf{Tier 3, $41.1 \pm 3.4$\% Coverage (Low)}: starting from a flat fabric, we grip at a random pick point and pull high in the air, drag in a random direction, and then drop, usually resulting in one or two corners hidden.
\end{itemize}

Figure~\ref{fig:fabric-start-states} shows examples of color and depth images of fabric initial states in simulation and real physical settings for all three tiers of difficulty. The supplementary material contains additional examples of images.

\section{Baseline Policies}\label{sec:baseline-policies}

We propose five baseline policies for fabric smoothing.
% which (except the random one) use the fabric state $\xi_t$ in simulation.

\subsubsection{Random}

As a naive baseline, we test a random policy that uniformly selects random pick points and pull directions.

\subsubsection{Highest (Max $z$)}

This policy, tested in Seita~et~al.~\cite{seita-bedmaking} grasps the highest point on the fabric. We get the pick point by determining $p$, the highest of the $25^2 = 625$ points from $\xi_t$. To compute the pull vector, we obtain the target coordinates by considering where $p$'s coordinates would be if the fabric is perfectly flat. The pull vector is then the vector from $p$'s current position to that target. Seita~et~al.~\cite{seita-bedmaking} showed that this policy can achieve reasonably high coverage, particularly when the highest point corresponds to a corner fold on the uppermost layer of the fabric.

\subsubsection{Wrinkle}

%One baseline method is to detect wrinkles and pull in the direction perpendicular to the largest wrinkle. Intuitively, this flattens the wrinkle, and the process can be repeated for subsequent (perhaps smaller) wrinkles.
Sun~et~al.~\cite{heuristic_wrinkles_2014} propose a two-stage algorithm to first identify wrinkles and then to derive a force parallel to the fabric plane to flatten the largest wrinkle. The process repeats for subsequent wrinkles. We implement this method by finding the point in the fabric of largest local height variance. Then, we find the neighboring point with the next largest height variance, treat the vector between the two points as the wrinkle, and pull perpendicular to it.

\subsubsection{Oracle}\label{ssec:expert-policy}

This policy uses complete state information from $\xi_t$ to find the fabric corner furthest from its fabric plane target, and pulls it towards that target. When a corner is occluded and underneath a fabric layer, this policy will grasp the point directly above it on the uppermost fabric layer, and the resulting pull usually decreases coverage.

\subsubsection{Oracle-Expose}

When a fabric corner is occluded, and other fabric corners are not at their targets, this policy picks above the hidden corner, but pulls away from the fabric plane target to reveal the corner for a subsequent action.

% Daniel: see these for details:
% https://github.com/BerkeleyAutomation/gym-fabric/pull/30 
% https://github.com/BerkeleyAutomation/gym-fabric/issues/28
% If we want to report on the number of transitions per episode, just do (num_transitions / num_episodes).
\begin{table}[t]
\caption{
\small
Results from the five baseline policies discussed in Section~\ref{sec:baseline-policies}. We report final coverage and the number of actions per episode. All statistics are from 2000 episodes, with tier-specific starting states. Both oracle policies (in bold) perform the best.
}
\centering
\begin{tabular}{l l | l l}
\textbf{Tier} & \textbf{Method} & \textbf{Coverage} & \textbf{Actions} \\ \hline 
1  & Random  &  25.0 +/- 14.6 & 2.43 +/- 2.2  \\
1  & Highest &  66.2 +/- 25.1 & 8.21 +/- 3.2  \\
1  & Wrinkle &  91.3 +/- 7.1  & 5.40 +/- 3.7  \\
1  & Oracle  &  \textbf{95.7 +/- 2.1}   & \textbf{1.76 +/- 0.8} \\ % \hline
1  & Oracle-Expose & \textbf{95.7 +/- 2.2} & \textbf{1.77 +/- 0.8}  \\ \hline
2  & Random  &  22.3 +/- 12.7  & 3.00 +/- 2.5  \\ 
2  & Highest &  57.3 +/- 13.0  & 9.97 +/- 0.3  \\
2  & Wrinkle &  87.0 +/- 10.8  & 7.64 +/- 2.8  \\
2  & Oracle  &  \textbf{94.5 +/- 5.4}   &  \textbf{4.01 +/- 2.0} \\  %\hline 
2  & Oracle-Expose & \textbf{94.6 +/- 5.0} & \textbf{4.07 +/- 2.2} \\  \hline
3  & Random  &  20.6 +/- 12.3  & 3.78 +/- 2.8  \\ 
3  & Highest &  36.3 +/- 16.3  & 7.89 +/- 3.2  \\ 
3  & Wrinkle &  73.6 +/- 19.0  & 8.94 +/- 2.0  \\ 
3  & Oracle  &  \textbf{95.1 +/- 2.3}  &  \textbf{4.63 +/- 1.1}  \\
3  & Oracle-Expose & \textbf{95.1 +/- 2.2} &  \textbf{4.70 +/- 1.1} \\
\end{tabular}
\vspace*{-10pt}
\label{tab:analytic}
\end{table}

% See Google Drive figure, named: traj_01_corners_perfect.
% Basically, run the oracle corner policy without the domain randomization. See the Google Drive for exact commands.
% UPDATE: actually there's an alternative, traj_02_corners_perfect. That one includes depth as well, and also shows the change in coverage at each time step. I think it is better for the paper so we can have that. See the exact command to reproduce in the Google Drive folder.
% I have commented out the caption that corresponds to the older figure (or you can also just get it from v1 of the paper on arXiv). Fortunately this time the second action also pulls over a fabric layer that is above the corner, so I don't have to change that part of the wording.
\begin{figure*}[t]
\center
\includegraphics[width=0.95\textwidth]{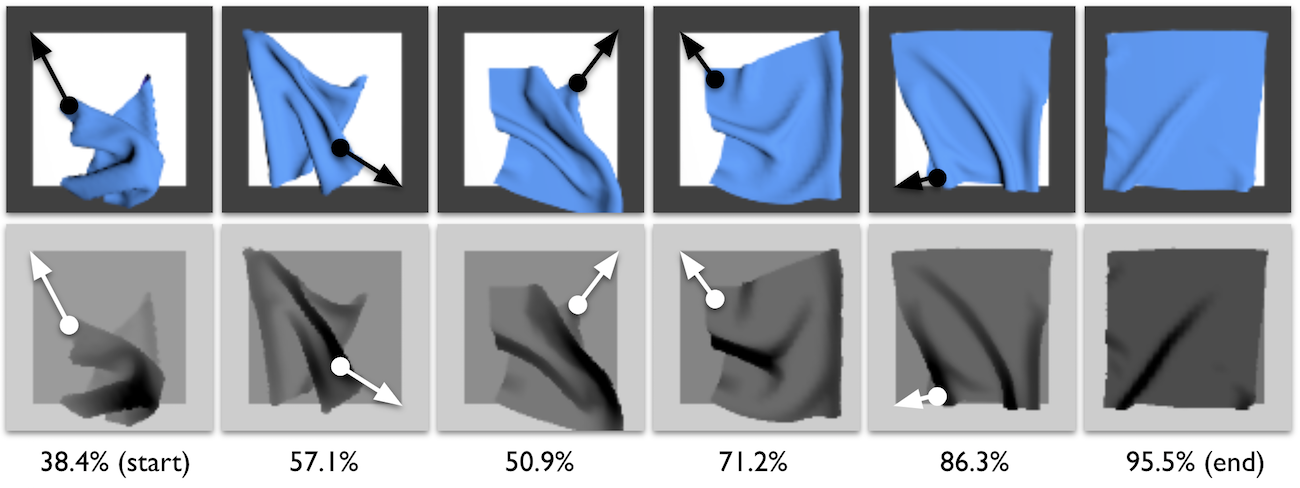}
\caption{
\small
%Example simulated episode of the oracle corner policy, from left to right. The policy uses the exact corner location and pulls the one furthest from its target on the white fabric plane.  Overlaid circles and arrows represent the action taken after the given state. The starting state (leftmost image) is drawn from Tier 3. In the second action, the fabric corner furthest from the target is slightly underneath the fabric, and the supervisor pulls at the fabric's top layer. Nonetheless, the subsequent pull (third image) is then able to reveal that fabric corner. The oracle policy took five actions before triggering the 92\% coverage threshold in the rightmost image.
Example simulated episode of the oracle corner supervisor policy, from left to right, drawn from a Tier 3 starting state with 38.4\% coverage. The policy uses the exact corner location from the fabric state (not the images) and pulls the one furthest from its target on the fabric plane. For visualization purposes, we show matching color and depth images from the episode without domain randomization. Overlaid circles and arrows represent the action taken after the given state. In the second action, the fabric corner furthest from the target is slightly underneath the fabric, and the supervisor pulls at the fabric's top layer. Nonetheless, the underlying corner is still moved closer to its target position, and subsequent pulls are able to achieve high coverage. The oracle policy took five actions before getting to 95.5\% coverage and triggering the 92\% threshold in the rightmost images.
}
\vspace*{-10pt}
\label{fig:corners-perfect}
\end{figure*}

\section{Simulation Results for Baseline Policies}\label{sec:sim-results-baselines}

We evaluate the five baseline fabric smoothing policies by running each for 2000 episodes in simulation. Each episode draws a randomized fabric starting state from one of three difficulty tiers (Section~\ref{ssec:starting-state}), and lasts for a maximum of 10 actions. Episodes can terminate earlier under two conditions: (1) if a pre-defined coverage threshold is obtained, or (2) the fabric is out of bounds over a certain threshold. For (1) we use 92\% as the threshold, which produces visually smooth fabric (e.g., see the last few images in Figure~\ref{fig:corners-perfect}) and avoids supervisor data being dominated by taking actions of short magnitudes at the end of its episodes. For (2) we define a fabric as out of bounds if it has any point which lies at least 25\% beyond the fabric plane relative to the full distance of the edge of the plane. This threshold allows the fabric to go slightly off the fabric plane which is sometimes unavoidable since a perfectly smoothed fabric is the same size as the plane. We do not allow a pick point to lie outside the plane.

Table~\ref{tab:analytic} indicates that both oracle policies attain nearly identical performance and have the highest coverage among the baseline policies, with about 95\% across all tiers. The wrinkles policy is the next best policy in simulation, with 91.3\%, 87.0\%, and 73.6\% final coverage for the three respective tiers, but requires substantially more actions per episode.

One reason why the oracle policy still performs well with occluded corners is that the resulting pulls can move those corners closer to their fabric plane targets, making it easier for subsequent actions to increase coverage. Figure~\ref{fig:corners-perfect} shows an example episode from the oracle policy on a tier 3 starting state. The second action pulls at the top layer of the fabric above the corner, but the resulting action still moves the occluded corner closer to its target.
%This allows subsequent actions to adjust the fabric to attain high coverage.

\section{Imitation Learning with DAgger}\label{sec:imitation}

We use the oracle (not oracle-expose) policy to generate supervisor data and corrective labels. For each tier, we generate 2000 episodes from the supervisor and use that as offline data. We train a fabric smoothing policy in simulation using imitation learning on synthetic images.  When behavior cloning on supervisor data, the robot's policy will learn the supervisor's actions on states in the training data, but generalize poorly outside the data distribution~\cite{pomerleau1989alvinn}. To address this, we use Dataset Aggregation (DAgger)~\cite{ross2011reduction}, which requests the supervisor to label the states the robot encounters when running its learned policy. A limitation of DAgger is the need for continued access to the supervisor's policy, rather than just offline data. During training, the oracle corner-pulling supervisor is able to efficiently provide an action to each data point encountered by accessing the underlying fabric state information, so in practice this does not cause problems.

\subsection{Policy Training Procedure}

We use domain randomization~\cite{domain_randomization} during training. For RGB images, we randomize the fabric color by selecting RGB values uniformly at random across intervals that include shades of blue, purple, pink, red, and gray. For RGB images, we additionally randomize the brightness with gamma corrections~\cite{Poynton:2003:DVH:640595} and vary the shading of the fabric plane. For depth (D) images, we make the images slightly darker to more closely match real depth images. For both RGB and D, we randomize the camera pose with independent Gaussian distributions for each of the position and orientation components. The supplementary material has examples of images after domain randomization.

The policy neural network architecture is similar to the one in Matas~et~al.~\cite{sim2real_deform_2018}. As input, the network consumes images with dimension $(100 \times 100 \times c)$, where the number of channels is $c=1$ for D, $c=3$ for RGB, or $c=4$ for RGBD images. It passes the input image through four convolutional layers, each with 32 filters of size $3\times 3$, followed by four dense layers of size 256 each, for a total of about 3.4 million parameters. The network produces a 4D vector with a hyperbolic tangent applied to make components within $[-1,1]$. We optimize using Adam with learning rate $10^{-4}$ and use $L_2$ regularization of $10^{-5}$. 

The imitation learning code uses OpenAI baselines~\cite{baselines} to make use of its parallel environment support. We run the fabric simulator in ten parallel environments, which helps to alleviate the major time bottleneck when training, and pool together samples in a shared dataset. We first train with a ``behavior cloning (BC) phase'' where we minimize the $L_2$ error on the offline supervisor data, and then use a ``DAgger phase'' which rolls out the agent's policy and applies DAgger. We use 500 epochs of behavior cloning based on when the network's $L_2$ error roughly converged on a held-out validation dataset. The DAgger phase runs until the agent collectively performs 50,000 total steps. Further training details are in the supplementary material.

\subsection{Simulation Experiments}

\begin{figure}[t]
\center
\includegraphics[width=0.48\textwidth]{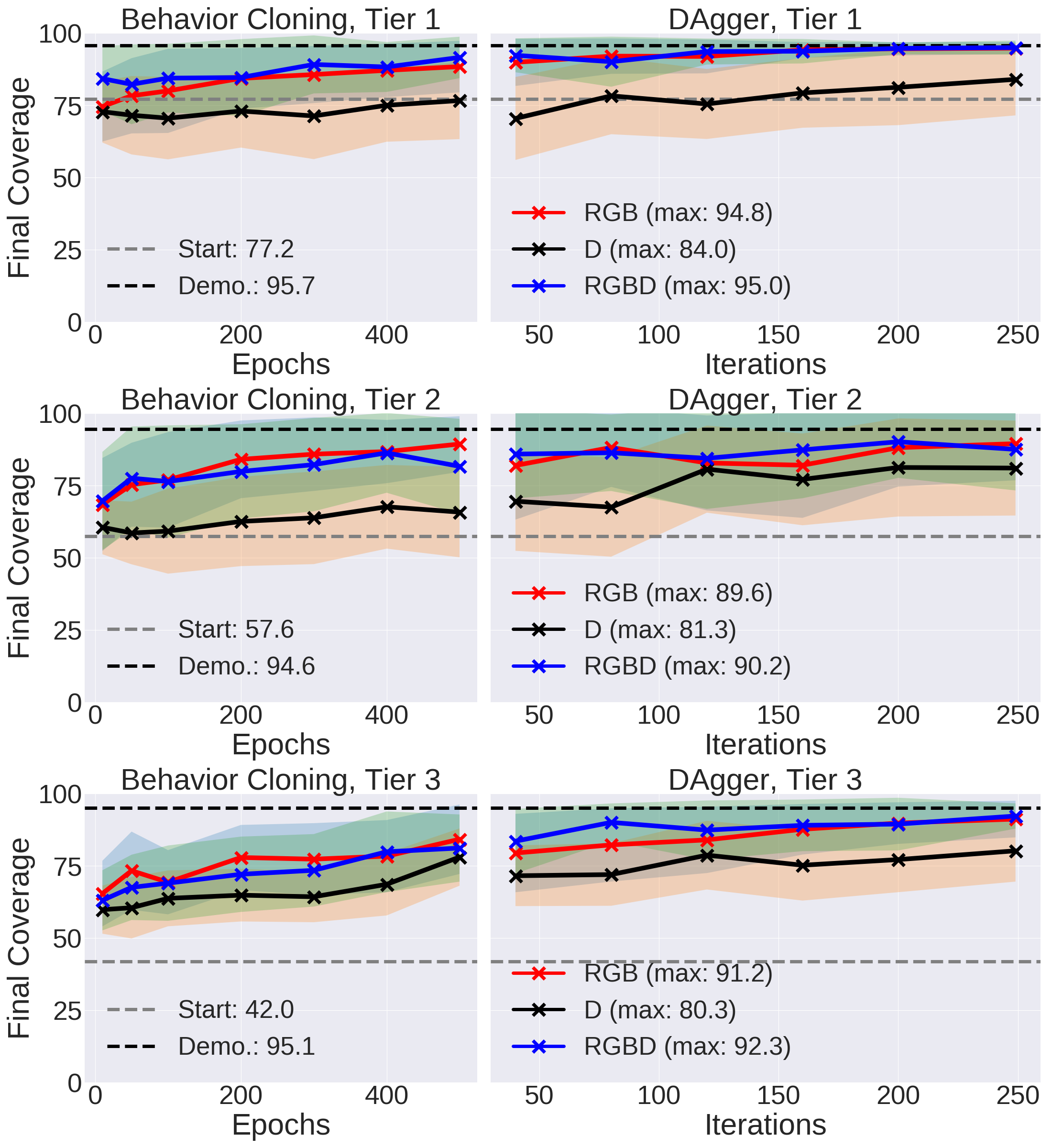}
\caption{
\small
Coverage over 50 simulated episodes at checkpoints (shown with ``X'') during behavior cloning (left) and DAgger (right), which begins right after the last behavior cloning epoch. Results, from top to bottom, are for tier 1, 2, and 3 starting states. We additionally annotate with dashed lines the average starting coverage and the supervisor's average final coverage. Results suggest that the RGB and RGBD policies attain strong coverage performance in simulation. Shaded regions represent one standard deviation range.
}
\vspace*{-10pt}
\label{fig:4-0-3}
\end{figure}

% ----------------------------------------
% Best overall coverage at any time in BC/DAgger:
% RGB:  avg([94.8, 89.6, 91.2])
% D:    avg([84.0, 81.3, 80.3])
% RGBD: avg([95.0, 92.3, 90.2])
%
% Best from the BC part:
% RGB:  ([88.4, 89.4, 84.2])
% D:    ([76.6, 67.7, 78.0])
% RGBD: ([91.6, 86.3, 81.2])
%
% Thus, the gains DUE TO DAGGER:
% RGB:  ([94.8-88.4, 89.6-89.4, 91.2-84.2])
% D:    ([84.0-76.6, 81.3-67.7, 80.3-78.0])
% RGBD: ([95.0-91.6, 92.3-86.3, 90.2-81.2])
%
% A bit clunky but here's the python code:
% (sum([94.8-88.4, 89.6-89.4, 91.2-84.2])/3  +  sum([84.0-76.6, 81.3-67.7, 80.3-78.0]) / 3 + sum([95.0-91.6, 92.3-86.3, 90.2-81.2]) / 3 ) /3
% Fortunately it's 6.1% which is THE SAME as I had earlier with color and depth only !! HA HA.
% ----------------------------------------
For all simulated training runs, we evaluate on 50 new tier-specific starting states that are not seen during training. Figure~\ref{fig:4-0-3} shows results across all tiers, suggesting that after behavior cloning, DAgger improves final coverage performance by 6.1\% when averaging over the nine experimental conditions (three image input modalities across three tiers). In addition, RGB policies attain better coverage in simulation than D policies with gains of 10.8\%, 8.3\%, and 10.9\% across respective tiers, which may be due to high color contrast between the fabric and fabric plane in the RGB images, as opposed to the D images (see Figure~\ref{fig:fabric-start-states}). The RGBD policies perform at a similar level as the RGB-based policies.

In all difficulty tiers, the RGB policies get higher final coverage performance than the wrinkles policy (from Table~\ref{tab:analytic}): 94.8\% over 91.3\%, 89.6\% over 87.0\%, and 91.2\% over 73.6\%, respectively, and gets close to the corner pulling supervisor despite only having access to image observations rather than underlying fabric state. Similarly, the RGBD policies outperform the wrinkles policy across all tiers. The depth-only policies outperform the wrinkles policy on tier 3, with 80.3\% versus 73.6\% coverage.

\section{Physical Experiments}\label{sec:physical-experiments}

% python scripts/combo_rollouts.py

The da Vinci Research Kit (dVRK) surgical robot~\cite{dvrk2014} is a cable-driven surgical robot with imprecision as reviewed in prior work~\cite{mahler2014case,seita_icra_2018}. We use a single arm with an end effector that can be opened to 75\degree, or a gripper width of 10mm. We set a fabric plane at a height and location that allows the end-effector to reach all points on it. To prevent potential damage to the grippers, the fabric plane is foam rubber, which allows us to liberally set the gripper height to be lower and avoids a source of height error present in~\cite{sim2real_deform_2018}. For the fabric, we cut a 5x5 inch piece from a Zwipes 735 Microfiber Towel Cleaning Cloth with a blue color within the distribution of domain randomized fabric colors. We mount a Zivid One Plus RGBD camera 0.9 meters above the workspace, which is used to obtain color and depth images.

\subsection{Physical Experiment Protocol} 

We manually create starting fabric states similar to those in simulation for all tiers. Given a starting fabric, we randomly run one of the RGB or D policies for one episode for at most 10 steps (as in simulation). Then, to make comparisons fair, we ``reset'' the fabric to be close to its starting state, and run the other policy. After these results, we then run the RGBD baseline to combine RGB and D images, again manipulating the fabric starting state to be similar among comparable episodes. To facilitate this process, we save all real images encountered and use them as a guide to creating the initial fabric configurations.

During preliminary trials, the dVRK gripper would sometimes miss the fabric by 1-2 mm, which is within the calibration error. To counter this, we measure structural similarity~\cite{ssim_2004} of the image before and after an action to check if the robot moved the fabric. If it did not, the next action is adjusted to be closer to the center of the fabric plane, and the process repeats until the robot touches fabric.

\subsection{Physical Experiment Results} 

\begin{figure*}[t]
\center
\includegraphics[width=1.0\textwidth]{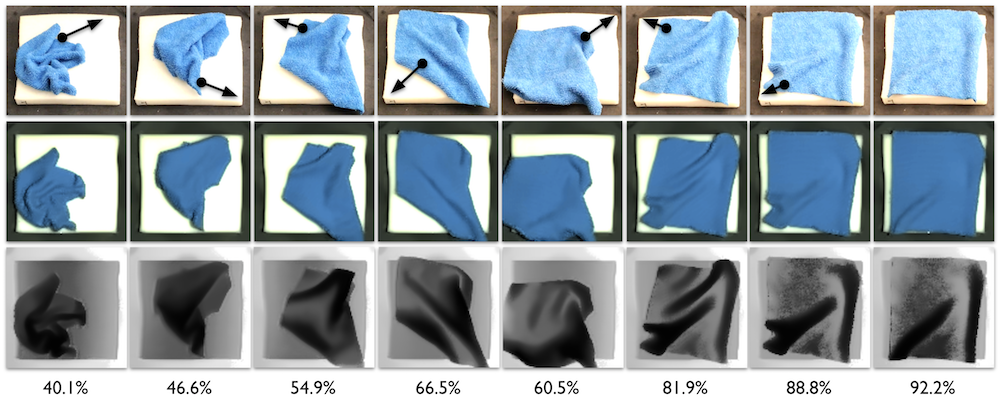}
\caption{
\small
% This is the older caption.
%An example episode (reproduced from Figure~\ref{fig:teaser}) taken by a learned policy trained on color images from tier 3 starting states. (Images are taken from the camera view used to record videos.) The leftmost image shows the starting state of the fabric, set to be highly wrinkled with at least the bottom right fabric corner hidden. The policy takes seven actions in this episode, with pick points and pull vectors indicated by the overlaid black arrows. Despite the highly wrinkled starting state, the policy is able to smooth fabric to get above 92\% coverage as shown at the rightmost image. 
An example episode taken by a learned policy trained on RGBD images from tier 3 starting states. The top row shows screen captures from the camera view used to record videos. The middle and bottom row show the processed RGB and D images that are passed into the neural network policy. The leftmost images show the starting state of the fabric, set to be highly wrinkled with at least the bottom left fabric corner hidden. The policy takes seven actions shown here, with pick points and pull vectors indicated by the overlaid black arrows in the top row. Despite the highly wrinkled starting state, along with hidden fabric corners (as shown in the first four columns) the policy is able to smooth fabric from 40.1\% to 92.2\% coverage as shown at the rightmost images.
}
\vspace*{-10pt}
\label{fig:trajectories}
\end{figure*}

% From https://github.com/BerkeleyAutomation/dvrk_python, run: `python analysis.py`. It automatically creates LaTeX-formatted tables. :-)
\begin{table}[t]
\caption{
\small
Physical experiments. We run 20 episodes for each of the tier 1 (T1), tier 2 (T2), and tier 3 (T3) fabric conditions, with RGB, D, and RGBD policies, for a total of $20  \times 3 \times 3 = 180$ episodes. We report: (1) starting coverage, (2) final coverage, with the highest values in each tier in bold, (3) maximum coverage at any point after the start state, and (4) the number of actions per episode. Results suggest that RGBD is the best policy with harder starting fabric configurations.
}
\centering
\begin{tabular}{l | l l l r }
& \textbf{(1) Start} & \textbf{(2) Final} & \textbf{(3) Max} & \textbf{(4) Actions} \\ \hline 
T1 RGB  & 78.4 +/- 4 & \textbf{96.2 +/- 2} & 96.2 +/- 2 & 1.8 +/- 1 \\
T1 D    & 77.9 +/- 4 & 78.8 +/- 24 & 90.0 +/- 10 & 5.5 +/- 4 \\
T1 RGBD & 72.5 +/- 4 & 95.0 +/- 2 & 95.0 +/- 2 & 2.1 +/- 1 \\ \hline
T2 RGB  & 58.5 +/- 6 & 87.7 +/- 13 & 92.7 +/- 4 & 6.3 +/- 3 \\
T2 D    & 58.7 +/- 5 & 64.9 +/- 20 & 85.7 +/- 8 & 8.3 +/- 3 \\
T2 RGBD & 55.0 +/- 5 & \textbf{91.3 +/- 8} & 92.7 +/- 6 & 6.8 +/- 3 \\ \hline
T3 RGB  & 46.2 +/- 4 & 75.0 +/- 18 & 79.9 +/- 14 & 8.7 +/- 2 \\
T3 D    & 47.0 +/- 3 & 63.2 +/- 9 & 74.7 +/- 10 & 10.0 +/- 0 \\
T3 RGBD & 41.7 +/- 2 & \textbf{83.0 +/- 10} & 85.8 +/- 6 & 8.8 +/- 2 \\ \hline
\end{tabular}
\vspace*{-10pt}
\label{tab:surgical}
\end{table}

We run 20 episodes for each combination of input modality (RGB, D, or RGBD) and tiers, resulting in 180 total episodes as presented in Table~\ref{tab:surgical}. We report starting coverage, ending coverage, maximum coverage across the episode after the initial state, and the number of actions. The maximum coverage allows for a more nuanced understanding of performance, because policies can take strong initial actions that achieve high coverage (e.g., above 80\%) but a single counter-productive action at the end can substantially lower coverage.

% Daniel: I used this to get the numbers:
%
%In [5]: sum([96.2-78.4, 87.7-58.5, 75.0-46.2]) / 3
%Out[5]: 25.266666666666666
%In [6]: sum([78.8-77.9, 64.9-58.7, 63.2-47.0]) / 3
%Out[6]: 7.766666666666666
%In [1]: sum([95.0-72.5, 91.3-55.0, 83.0-41.7]) / 3
%Out[1]: 33.36666666666667
%
% For final coverage that we get, it's similar as above, except we don't subtract anything. I.e.:
%
%In [7]: sum([ 96.2 , 87.7 , 75.0 ] ) / 3
%Out[7]: 86.3
%In [8]: sum([ 78.8 , 64.9 , 63.2 ] ) / 3
%Out[8]: 68.96666666666665
%In [1]: sum( [95.0 , 91.3, 83.0 ] ) / 3
%Out[1]: 89.76666666666667
%
% These numbers will probably be good enough to report.
Results suggest that, despite not being trained on real images, the learned policies can smooth physical fabric. All policies improve over the starting coverage across all tiers. Final coverage averaged across all tiers is 86.3\%, 69.0\%, and 89.8\% for RGB, D, and RGBD policies, respectively, with net coverage gains of 25.2\%, 7.8\%, and 33.4\% over starting coverage. In addition, the RGB and RGBD policies deployed on tier 1 starting states each achieve the 92\% coverage threshold 20 out of 20 times. While RGB has higher final coverage on tier 1 starting states, the RGBD policies appear to have stronger performance on the more difficult starting states without taking considerably more actions.

Qualitatively, the RGB and RGBD-trained policies are effective at ``fine-tuning'' by taking several short pulls to trigger at least 92\% coverage. For example, Figure~\ref{fig:trajectories} shows an episode taken by the RGBD policy trained on tier 3 starting states. It is able to smooth the highly wrinkled fabric despite several corners hidden underneath fabric layers. The depth-only policies do not perform as well, but this is in large part because the depth policy sometimes takes counterproductive actions after several reasonable actions. This may be in part due to uneven texture on the fabric we use, which is difficult to replicate in simulated depth images.

%The supplementary material contains videos of every tier and policy combination.

% \subsection{Other Shapes?}
% 
% \bnote{Daniel: might be interesting to see how well the physical policies perform on different shape fabrics?}

\subsection{Failure Cases}

\begin{figure}[t]
\center
% We had this v02 in the ICRA submission.
%\includegraphics[width=0.45\textwidth]{potential_failure_v02_color_correction.png}
\includegraphics[width=0.45\textwidth]{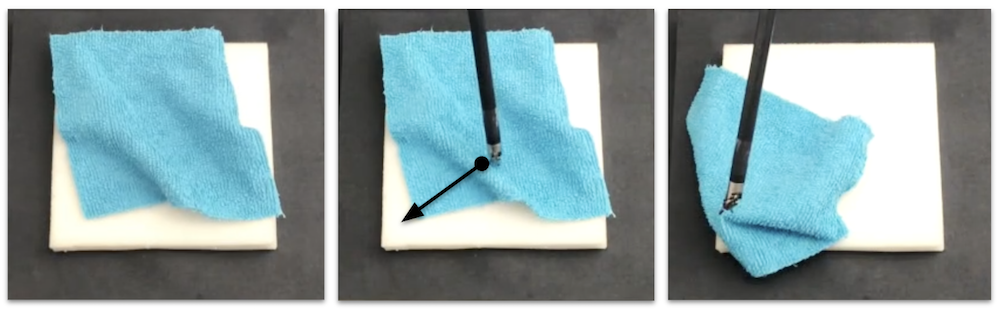}
\caption{
\small
A poor action from a policy trained on only depth images. Given the situation shown in the left image, the policy picks a point near the center of the fabric. The resulting pick, followed by the pull to the lower left, causes a major decrease in coverage and may make it hard for the robot to recover.
}
\vspace*{-10pt}
\label{fig:failure-case}
\end{figure}

The policies are sometimes susceptible to performing highly counter-productive actions. In particular, the depth-only policies can fail by pulling near the center of the fabric for fabrics that are already nearly smooth, as shown in Figure~\ref{fig:failure-case}. This results in poor coverage and may lead to cascading errors where one poor action can lead fabric to reach states that are not seen in training. 

One cause may be that there are several fabric corners that are equally far from their targets, which creates ambiguity in which corner should be pulled. One approach to mitigate this issue, in addition to augmenting depth with RGB-based data as in the RGBD policies we train, is to formulate corner picking with a mixture model to resolve this ambiguity.

%===============================================================================

\section{Conclusion and Future Work}\label{sec:conclusions}

We investigate baseline and learned policies for fabric smoothing. Using a low fidelity fabric simulator and a custom environment, we train policies in simulation using DAgger with a corner pulling supervisor. We use domain randomization to transfer policies to a surgical robot. When testing on fabric of similar color to those used in training, RGB-based and RGBD-based policies achieve higher coverage than D-based policies. On the harder starting fabric configurations, the combined RGBD-based policies get the highest coverage among the tested policies, suggesting that depth is beneficial.

In future work, we will test on fabric shapes and configurations where corner pulling policies may get poor coverage. We plan to apply state-of-the-art deep reinforcement learning methods using the simulation environment to potentially learn richer policies that can explicitly reason over multiple time steps and varying geometries. To improve sample efficiency, we will consider model-based approaches such as Visual Foresight~\cite{visual_foresight_2018} and future image similarity~\cite{model_based_bc_2019}. We will utilize higher-fidelity fabric simulators such as ARCSim~\cite{arcsim2012}. Finally, we would like to extend the method beyond fabric coverage to tasks such as folding and wrapping, and will apply it to ropes, strings, and other deformable objects.

% Daniel: for the ICRA 2019 submission (which got rejected), we got an angry comment from the "meta review" who said using tiny font for acknowledgments was unacceptable. Fine, let's keep it at small BUT DO NOT MAKE IT ANY SMALLER!!!
\section*{Acknowledgments}
{\small
This research was performed at the AUTOLAB at UC Berkeley in affiliation with Honda Research Institute USA, the Berkeley AI Research (BAIR) Lab, Berkeley Deep Drive (BDD), the Real-Time Intelligent Secure Execution (RISE) Lab, and the CITRIS ``People and Robots'' (CPAR) Initiative, and by the Scalable Collaborative Human-Robot Learning (SCHooL) Project, NSF National Robotics Initiative Award 1734633. The authors were supported in part by Siemens, Google, Amazon Robotics, Toyota Research Institute, Autodesk, ABB, Samsung, Knapp, Loccioni, Intel, Comcast, Cisco, Hewlett-Packard, PhotoNeo, NVidia, and Intuitive Surgical. Daniel Seita is supported by a National Physical Science Consortium Fellowship. We thank Jackson Chui, Michael Danielczuk, Shivin Devgon, and Mark Theis.
}

\bibliographystyle{IEEEtranS}
\bibliography{example}

\cleardoublepage
\appendices

We structure this appendix as follows:

\begin{itemize}
    \item Appendix~\ref{app:fabric} provides details on the fabric simulator.
    \item Appendix~\ref{app:baselines} discusses the baseline pick point methods from Section~\ref{sec:baseline-policies}.
    \item Appendix~\ref{app:imitation} explains the imitation learning pipeline, and includes discussion about the neural network architecture and the domain randomization.
    \item Appendix~\ref{app:experiment-details} describes the experiment setup with the da Vinci Surgical robot.
    \item Appendix~\ref{app:more-experiments} presents additional experiments with yellow fabric.
\end{itemize}

\section{Fabric Environment}\label{app:fabric}

The fabric simulator is implemented in Python with Cython for increased speed. The simulator is low fidelity compared to more accurate simulators such as ARCSim~\cite{arcsim2012}, but has the advantage of being easier to adapt to the smoothing task we consider and enabling massive data generation. Some relevant hyperparameters are shown in Table~\ref{tab:fabric}.

\subsection{Actions}\label{app:grasping}

The fabric smoothing environment is implemented with a standard OpenAI gym~\cite{gym} interface. Each action is broken up into six stages: (1) a grasp, (2) a pull up, (3) a pause, (4) a linear pull towards a target, (5) a second pause, and (6) a drop. Steps (2) through (6) involve some number of iterations, where each iteration changes the coordinates of ``pinned'' points on $\xi_t$ and then calls one ``update'' method for the fabric simulator to adjust the other, non-pinned points.

\subsubsection{Grasp}

We implement a grasp by first simulating a gripper moving downwards from a height higher than the highest fabric point, which simulates grasping only the top layer of the fabric. At a given height $z_t$, to decide which points in $\xi_t$ are grasped given pick point $(x_t,y_t)$, we use a small sphere centered at $(x_t,y_t,z_t)$ with a radius of 0.003 units, where units are scaled so 1 represents the length of a side of the fabric plane. If no points are gripped, then we lower $z_t$ until a point is within the radius. In practice, this means usually 2-5 out of the $25^2=625$ points on the cloth are grasped for a given pick point.  Once any of the fabric's points are within the gripper's radius, those points are considered fixed, or ``pinned''.

\subsubsection{Pull Up} For 50 iterations, the pull adjusts the $z$-coordinate of any pinned point by $dz = 0.0025$ units, and keeps their $x$ and $y$ coordinates fixed. In practice, tuning $dz$ is important. If it is too low, an excessive amount of fabric-fabric collisions can happen, but if it is too high, then coverage substantially decreases. In future work, we will consider dynamically adjusting the height change so that it is lower if current fabric coverage is high.

\subsubsection{First Pause} For 80 iterations, the simulator keeps the pinned points fixed, and lets the non-pinned points settle.

\subsubsection{Linear Pull to Target} To implement the pull, we adjust the $x$ and $y$ coordinates of all pinned points by a small amount each time step (leaving their $z$ coordinates fixed), in accordance with the deltas $(\Delta x_t, \Delta y_t)$ in the action. The simulator updates the position of the non-pinned points based on the implemented physics model. This step is run for a variable amount of iterations based on the pull length.

\subsubsection{Second Pause} For 300 iterations, the simulator keeps the pinned points fixed, and lets the non-pinned points settle. This period is longer than the first pause because normally more non-pinned points are moving after a linear pull to the target compared to a pull upwards.

\subsubsection{Drop} Finally, the pinned points are ``un-pinned'' and thus are allowed to lower due to gravity. For 1000 iterations, the simulator lets the entire cloth settle and stabilize for the next action.

\begin{table}[t]
\caption{
\small
Fabric simulator hyperparameters. The spring constant $k_s$ is in Equation~\ref{eq:hookes} and damping $d$ is in Equation~\ref{eq:final}.
}
\centering
\begin{tabular}{l r}
\textbf{Hyperparameter} & \textbf{Value} \\ \hline
Number of Points      & $25\times 25 = 625$ \\
Damping $d$           & 0.020 \\
Spring Constant $k_s$  & 5000.0 \\
Self-collision thickness & 0.020 \\
Height change $dz$ per iteration  & 0.0025 \\
\end{tabular}
%\vspace*{-10pt}
\label{tab:fabric}
\end{table}

\subsection{Starting State Distributions}

We provide details on how we generate starting states from the three distributions we use (see Section~\ref{ssec:starting-state}).

\begin{itemize}
\item \textbf{Tier 1}. We perform a sequence of two pulls with pick point randomly chosen on the fabric, pull direction randomly chosen, and pull length constrained to be short (about 10-20\% of the length of the fabric plane). If coverage remains above 90\% after these two pulls we perform a third short (random) pull. 
\item \textbf{Tier 2}. For this tier only, we initialize the fabric in a vertical orientation with tiny noise in the direction perpendicular to the plane of the fabric. Thus the first action in Tier 2 initialization is a vertical drop over one of two edges of the plane (randomly chosen). We then randomly pick one of the two corners at the top of the dropped fabric and drag it approximately toward the center for about half of the length of the plane. Finally we grip a nearby point and drag it over the exposed corner in an attempt to occlude it, again pulling for about half the length of the plane.
\item \textbf{Tier 3}. Initialization consists of just one high pull. We choose a random pick point, lift it about 4-5 times as high as compared to a normal action, pull in a random direction for 10-25\% of the length of the plane, and let the fabric fall, which usually creates less coverage and occluded fabric corners.
\end{itemize}

The process induces a distribution over starting states, so the agent never sees the same starting fabric state.

These starting state distributions do not generally produce a setting when we have a single corner fold that is visible on top of the fabric, as that case was frequently shown and reasonably approached in prior work~\cite{seita-bedmaking}.

\section{Details on Baseline Policies}\label{app:baselines}

We describe the implementation of the analytic methods from Section~\ref{sec:baseline-policies} in more detail.

\subsection{Highest (Max $z$)}

We implement this method by using the underlying state representation of the fabric $\xi_t$, and not the images $\bo_t$. For each $\xi_t$, we iterate through all fabric point masses and obtain the one with the highest z-coordinate value. This provides the pick point. For the pull vector, we deduce it from the location of the fabric plane where it would be located if the fabric were perfectly flat.

To avoid potentially getting stuck repeatedly pulling the same point (which happens if the pull length is very short and the fabric ends up ``resetting'' to the prior state), we select the five highest points on the fabric, and then randomize the one to pick. This policy was able to achieve reasonable coverage performance in Seita~et~al.~\cite{seita-bedmaking}.

\subsection{Wrinkles}

The implementation approximates the method in Sun~et~al.~\cite{heuristic_wrinkles_2014}; implementing the full algorithm is difficult due to its complex, multi-stage nature and the lack of open source code. Their wrinkle detection method involves computing variance in height in small neighborhoods around each pixel in the observation image $\bo_t$, $k$-means clustering on the pixels with average variance above a certain threshold value, and hierarchical clustering on the clusters found by $k$-means to obtain the largest wrinkle. We approximate their wrinkle detection method by isolating the point of largest local variance in height using $\xi_t$. Empirically, this is accurate at selecting the largest wrinkle. To estimate wrinkle direction, we find the neighboring point with the next largest variance in height.

We then pull perpendicular to the wrinkle direction. Where Sun~et~al.~\cite{heuristic_wrinkles_2014} constrains the perpendicular angle to be one of eight cardinal directions (north, northeast, east, southeast, south, southwest, west, northwest), we find the exact perpendicular line and its two intersections with the edges of the fabric. We choose the closer of these two as the pick point and pull to the edge of the plane.

\subsection{Oracle}

For the oracle policy, we assume we can query the four corners of the fabric and know their coordinates. Since we know the four corners, we know which of the fabric plane corners (i.e., the targets) to pull to. We pick the pull based on whichever fabric corner is furthest from its target.

%This method performs reliably well if all the fabric corners are exposed. If a fabric corner is covered by a fold, this method may peform poorly as the action parameterization would cause the grip to grip at the top layer of the fabric over the fold, hence it is a ``soft'' upper bound on performance. %In practice, physical robot would not be able to get these exact points in general.

\subsection{Oracle Expose}

The oracle expose policy is an extension to the oracle policy. In addition to knowing the exact position of all corners, the policy is also aware of the occlusion state of all corners. The occlusion state is a 4D boolean vector which indicates a 1 if a given corner is visible to the camera from the top-down view or a 0 if it is occluded. The oracle expose policy will try to solve all visible corners in a similar manner to the oracle policy using its extended state knowledge. If all four corners are occluded, or all visible corners are within a threshold of their target positions, the oracle expose policy will perform a revealing action on an occluded corner. We implement the revealing action as a fixed length pull 180 degrees away from the angle to the target position. This process is repeated until the threshold coverage is achieved.

\section{Details on Imitation Learning}\label{app:imitation}

\subsection{DAgger Pipeline}

\begin{table}[t]
\caption{
\small
Hyperparameters for the main DAgger experiments. 
}
\centering
\begin{tabular}{l r}
\textbf{Hyperparameter} & \textbf{Value} \\  \hline
Parallel environments & 10 \\
Steps per env, between gradient updates & 20 \\
Gradient updates after parallel steps & 240 \\
Minibatch size            & 128 \\
%Number of demo. samples & Varies by tier \\
Supervisor (offline) episodes & 2000 \\
Policy learning rate   & 1e-4 \\
Policy $L_2$ regularization parameter  & 1e-5 \\
Behavior Cloning epochs & 500 \\
DAgger steps after Behavior Cloning    & 50000 \\
%Discount factor $\gamma$ & 0.99 \\
\end{tabular}
\vspace*{-10pt}
\label{tab:learning-hyperparams}
\end{table}

%We collect supervisor data by running the oracle corner policy (not oracle expose) for 2000 trajectories for each of the three starting state tiers. We then run behavior cloning on this offline data for 500 epochs before running DAgger.

Each DAgger ``iteration'' rolls out 10 parallel environments for 20 steps each (hence, 200 total new samples) which are labeled by the oracle corner policy. These are added to a growing dataset of samples which includes the supervisor's original offline data. After 20 steps per parallel environment, we draw 240 minibatches of size 128 each for training. Then the process repeats with the agent rolling out its new policy. DAgger hyperparameters are in Table~\ref{tab:learning-hyperparams}. In practice, the $L_2$ regularization for the policy impacts performance significantly. We use $10^{-5}$ as using $10^{-4}$ or $10^{-3}$ leads to substantially worse performance. We limit the total number of DAgger steps to 50,000 due to compute and time limitations; training for substantially more steps is likely to yield further improvements.

As described in Section~\ref{sec:imitation}, the policy neural network architecture has four convolutional layers, each with 32 filters of size $3\times 3$, followed by dense layers of size 256 each, for a total of about 3.4 million parameters. The parameters and their dimensions, assuming the input is RGB (with three channels) and ignoring biases for simplicity, are:

\footnotesize
\begin{verbatim}
policy/convnet/c1   864 params (3, 3, 3, 32)
policy/convnet/c2   9216 params (3, 3, 32, 32)
policy/convnet/c3   9216 params (3, 3, 32, 32)
policy/convnet/c4   9216 params (3, 3, 32, 32)
policy/fcnet/fc1    3276800 params (12800, 256)
policy/fcnet/fc2    65536 params (256, 256)
policy/fcnet/fc3    65536 params (256, 256)
policy/fcnet/fc4    1024 params (256, 4)
Total model parameters: 3.44 million
\end{verbatim}
\normalsize

% don't want this figure on its own!
\begin{figure}[t]
\center
\includegraphics[width=0.48\textwidth]{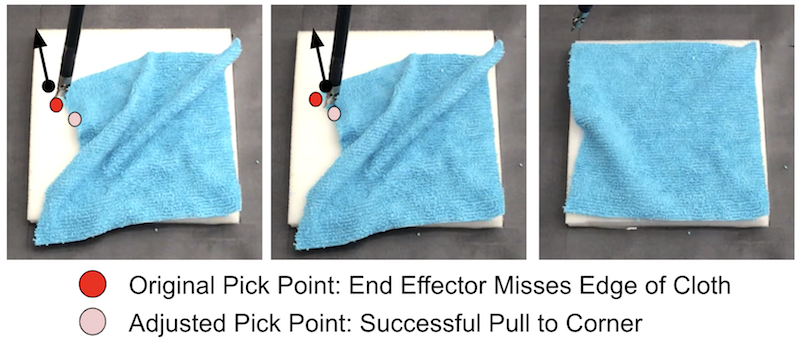}
\caption{
\small
Correcting the dVRK when it slightly misses fabric. Left: the dVRK executes a pick point (indicated with the red circle) but barely misses the fabric. Middle: it detects that the fabric has not changed, and the resulting action is constrained to be closer to the center and touches the fabric (light pink circle). Right: the pull vector results in high coverage.
}
%\vspace*{-10pt}
\label{fig:misses_cloth}
\end{figure}

\subsection{Domain Randomization and Simulated Images}

\begin{figure*}[t]
\center
\includegraphics[width=0.98\textwidth]{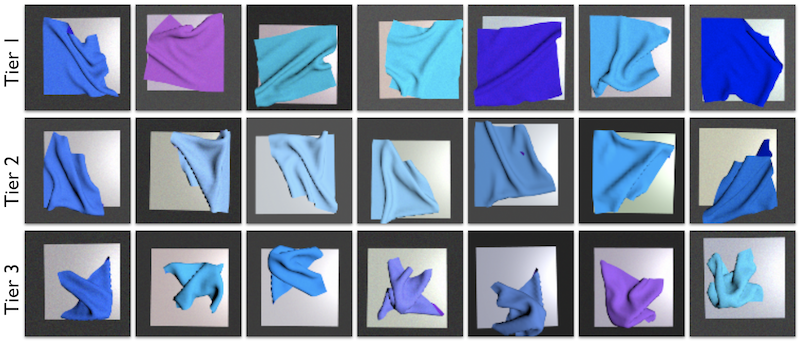}
\caption{
\small
Representative simulated color images of examples of starting fabric states drawn from the distributions specified in Section~\ref{ssec:starting-state}. All images are of dimension $100\times 100 \times 3$. Top row: tier 1. Middle row: tier 2. Bottom row: tier 3.  Domain randomization is applied on the fabric color, the shading of the white background plane, the camera pose, and the overall image brightness, and then we apply uniform random noise to each pixel.
}
%\vspace*{-10pt}
\label{fig:fabric-rgb}
\end{figure*}

\begin{figure*}[t]
\center
\includegraphics[width=0.98\textwidth]{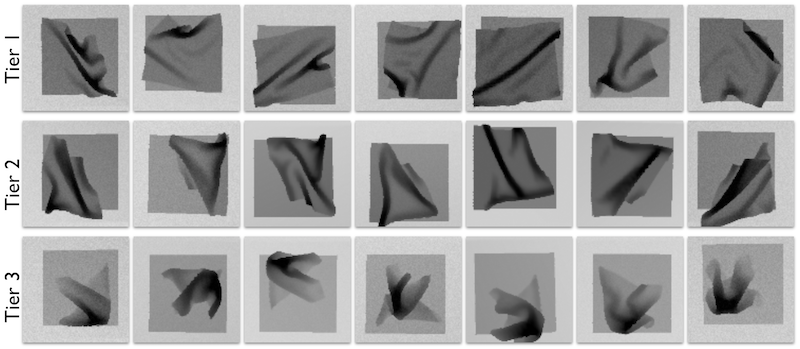}
\caption{
\small
Representative simulated depth images of examples of starting fabric states drawn from the distributions specified in Section~\ref{ssec:starting-state}, shown in a similar manner as in Figure~\ref{fig:fabric-rgb}. Images are of dimension $(100\times 100)$. Domain randomization is applied on the camera pose and the image brightness, and then we apply uniform random noise to each pixel.
}
%\vspace*{-10pt}
\label{fig:fabric-depth}
\end{figure*}

To transfer a policy to a physical robot, we use domain randomization~\cite{domain_randomization} on color (RGB) and depth (D) images during training. We randomize fabric colors, the shading of the fabric plane, and camera pose. We do not randomize the simulator's parameters.%as done in OpenAI~et~al.~\cite{openai-dactyl} and leave this to future work.

Figures~\ref{fig:fabric-rgb} and~\ref{fig:fabric-depth} show examples of simulated images with domain randomization applied. We specifically applied the following randomization, in order:

\begin{itemize}
\item For RGB images, we apply color randomization. The fabric background and foreground colors are set at default RGB values of $[0.07, 0.30, 0.90]$ and $[0.07, 0.05, 0.60]$, respectively, creating a default blue color. With domain randomization, we create a random noise vector of size three where each component is independently drawn from ${\rm Unif}[-0.35, 0.35]$ and then add it to both the background and foreground colors. Empirically, this creates images of various shades ``centered'' at the default blue value.
\item For both RGB and D images, we apply camera pose randomization, with Gaussian noise added independently to the six components of the pose (three for position using meters, three for orientation using degrees). Gaussians are drawn centered at zero with standard deviation 0.04 for positions and 0.9 for degrees.
\item After Blender produces the image, we next adjust the brightness of RGB images via OpenCV gamma corrections\footnote{\url{https://www.pyimagesearch.com/2015/10/05/opencv-gamma-correction/}}, with $\gamma = 1$ representing no brightness change. We draw $\gamma \sim {\rm Unif}[0.8, 1.2]$ for RGB images. To make simulated depth images darker and more closely match the real images, we draw a random value $\beta \sim {\rm Unif}[-50,-40]$ and add $\beta$ to each pixel.
\end{itemize}

Only after the above are applied, do we then independently add uniform noise to each pixel. For each full image with pixel values between 0 and 255, we draw a uniform random variable $l \sim {\rm Unif}[-15,15]$ between -15 and 15. We then draw additive noise $\epsilon \sim {\rm Unif}[-l,l]$ for each pixel independently.

\section{Experiment Setup Details}\label{app:experiment-details}

\subsection{Image Processing Pipeline}

\begin{figure*}[t]
\center
\includegraphics[width=1.00\textwidth]{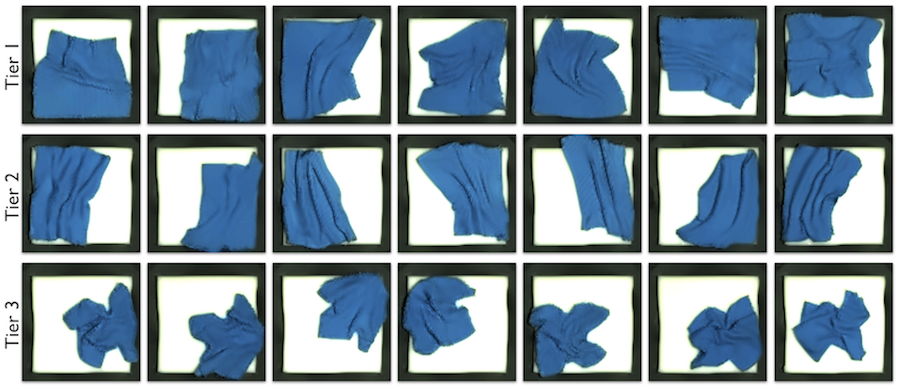}
\caption{
\small
Representative examples of color images from the physical camera used for the surgical robot experiments, so there is no domain randomization applied here. We manipulated fabrics so that they appeared similar to the simulated states. The color images here correspond to the depth images in Figure~\ref{fig:real-fabric-depth}.
}
%\vspace*{-10pt}
\label{fig:real-fabric-rgb}
\end{figure*}

\begin{figure*}[t]
\center
\includegraphics[width=1.00\textwidth]{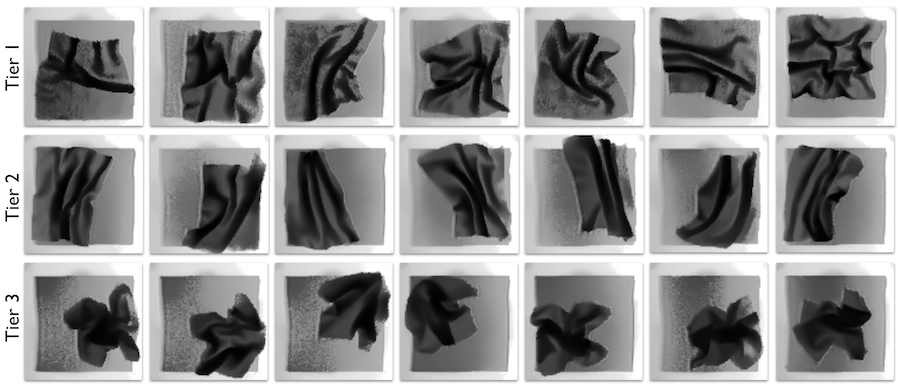}
\caption{
\small
Representative examples of depth images from the physical camera used for the surgical robot experiments, so there is no domain randomization applied here. We manipulated fabrics so that they appeared similar to the simulated states. The depth images here correspond to the color images in Figure~\ref{fig:real-fabric-rgb}.
}
%\vspace*{-10pt}
\label{fig:real-fabric-depth}
\end{figure*}

The original color and depth images come from the mounted Zivid One Plus RGBD camera, and are processed in the following ordering:

\begin{itemize}
    \item For depth images only, we apply in-painting to fill in missing values (represented as ``NaN''s) in depth images based on surrounding pixel values.
    \item Color and depth images are then cropped to be $100\times 100$ images that allow the entire fabric plane to be visible, along with some extra background area.
    \item For depth images only, we clip values to be within a minimum and maximum depth range, tuned to provide depth images that looked reasonably similar to ones processed in simulation. We convert images to three channels by triplicating values across the channels. We then scale pixel values to be within $[0,255]$ and apply the OpenCV equalize histogram function for all three channels.
    \item For depth and color images, we apply bilateral filtering and then de-noising, both implemented using OpenCV functions. These help smooth the uneven fabric texture without sacrificing cues from the corners.
\end{itemize}

Figures~\ref{fig:real-fabric-rgb} and~\ref{fig:real-fabric-depth} show examples of real fabric images that policies take as input, after processing.  These are then passed as input to the policy neural network.

\subsection{Physical Experiment Setup and Procedures}

To map from neural network output to a position with respect to the robot's frame, we calibrate the positions by using a checkerboard on top of the fabric plane. We move the robot's end effectors with the gripper facing down to each corner of the checkerboard and record positions. During deployment, for a given coordinate frame, we perform bilinear interpolation to figure out the robot position from the four surrounding known points. After calibration, the robot reached positions on the fabric plane to 1-2 mm of error. Figure~\ref{fig:misses_cloth} shows a visualization of the heuristic we employ to get the robot to grasp fabric when it originally misses by 1-2mm.

\section{Additional Experiments}\label{app:more-experiments}

%\subsection{Experiments With Yellow Fabrics}

\begin{figure}[t]
\center
\includegraphics[width=0.45\textwidth]{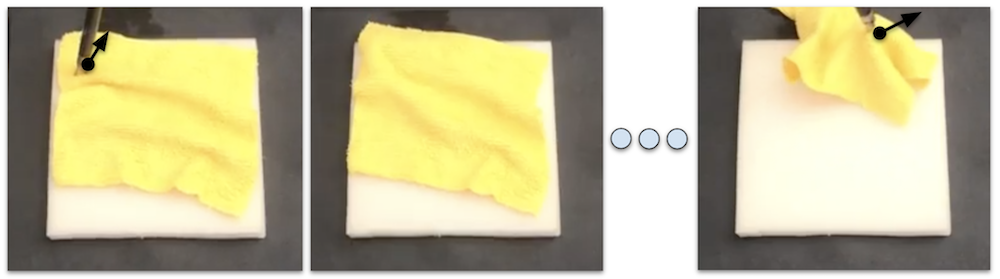}
\caption{
\small
An example episode taken by a tier 1, RGB-trained policy on yellow fabric. The first action (left image) picks to the upper left and pulls the fabric away from the plane (black arrow). The process repeats for several actions and, as shown in the last image, the fabric barely covers the plane.
}
%\vspace*{-10pt}
\label{fig:failure-yellow}
\end{figure}

% From https://github.com/BerkeleyAutomation/dvrk_python, run: `python analysis.py`. It automatically creates LaTeX-formatted tables. :-)
\begin{table}[t]
\caption{
\small
Physical experiments for 5 trajectories with yellow fabric. We report the same statistics as in Table~\ref{tab:surgical} for the two same policies (T1 RGB and T1 D) trained on tier 1 starting states.
}
\centering
\begin{tabular}{l | l l l r}
& \textbf{(1) Start} & \textbf{(2) Final} & \textbf{(3) Max} & \textbf{(4) Actions} \\ \hline 
T1 RGB & 81.5 +/- 3 & 71.7 +/- 25 & 89.6 +/- 6  & 7.6 +/- 2 \\
T1 D & 83.1 +/- 2 & \textbf{85.9 +/- 15} & \textbf{91.9 +/- 5} & \textbf{4.6 +/- 4} \\
\end{tabular}
%\vspace*{-10pt}
\label{tab:surgical-other-fabrics}
\end{table}

To further test RGB versus D policies, we use the same two policies trained on tier 1 starting states and deploy them on yellow fabric. The color distribution ``covered'' by domain randomization includes shades of blue, purple, pink, red, and gray, but not yellow. We recreate five starting fabric conditions where, with a blue fabric, the RGB policy attained at least 92\% coverage in just one action.

Results in Table~\ref{tab:surgical-other-fabrics} indicate poor performance from the RGB policy, as coverage decreases from 81.5\% to 71.7\%. Only two out of five trajectories result in at least 92\% coverage. We observe behavior shown in Figure~\ref{fig:failure-yellow} where the policy fails to pick at a corner or to pull in the correct direction. The depth-only policy is invariant to colors, and is able to achieve higher ending coverage of 85.9\%. This is higher than the 78.8\% coverage reported in Table~\ref{tab:surgical} due to relatively easier starting states.

\end{document}